\documentclass{article}

\usepackage{microtype}
\usepackage{graphicx}
\usepackage{subfigure}
\usepackage{booktabs} 

\usepackage{hyperref}

\usepackage[accepted]{icml2023}

\usepackage{amsmath}
\usepackage{amssymb}
\usepackage{mathtools}
\usepackage{amsthm}
\usepackage{todonotes}
\usepackage[capitalize,noabbrev]{cleveref}

\theoremstyle{plain}
\newtheorem{theorem}{Theorem}[section]

\theoremstyle{definition}
\newtheorem{definition}[theorem]{Definition}

\theoremstyle{remark}

\usepackage{makecell}
\usepackage[export]{adjustbox} 

\icmltitlerunning{Removing Model Behaviors with Targeted Ablation}
\def\eqn#1{\begin{equation}#1\end{equation}}
\def\agn#1{\begin{align}#1\end{align}}
\def\Ls{\mathcal{L}}
\def\md{\mathcal{M}}
\def\Db{\mathcal{D}}
\usepackage{stfloats}
\usepackage{caption}
\begin{document}

\twocolumn[
\icmltitle{Circuit Breaking: Removing Model Behaviors with Targeted Ablation}



\icmlsetsymbol{equal}{*}

\begin{icmlauthorlist}
\icmlauthor{Maximilian Li}{equal,yyy}
\icmlauthor{Xander Davies}{equal,yyy}
\icmlauthor{Max Nadeau}{equal,yyy}
\end{icmlauthorlist}

\icmlaffiliation{yyy}{Harvard University}

\icmlcorrespondingauthor{Maximilian Li}{maxli@college.harvard.edu}
\icmlcorrespondingauthor{Xander Davies}{xanderlaserdavies@gmail.com}
\icmlcorrespondingauthor{Max Nadeau}{nadeau.max@gmail.com}


\vskip 0.3in
]



\printAffiliationsAndNotice{\icmlEqualContribution} 

\begin{abstract}
     Language models often exhibit behaviors that improve performance on a pre-training objective but harm performance on downstream tasks. We propose a novel approach to removing undesirable behaviors by ablating a small number of causal pathways between model components, with the intention of disabling the computational circuit responsible for the bad behavior. Given a small dataset of inputs where the model behaves poorly, we learn to ablate a small number of important causal pathways. In the setting of reducing GPT-2 toxic language generation, we find ablating just 12 of the 11.6K causal edges mitigates toxic generation with minimal degradation of performance on other inputs.
     
    \vspace*{-2em}

\end{abstract}

\begin{figure*}[!b]
    \centering
    \includegraphics[width=6in]{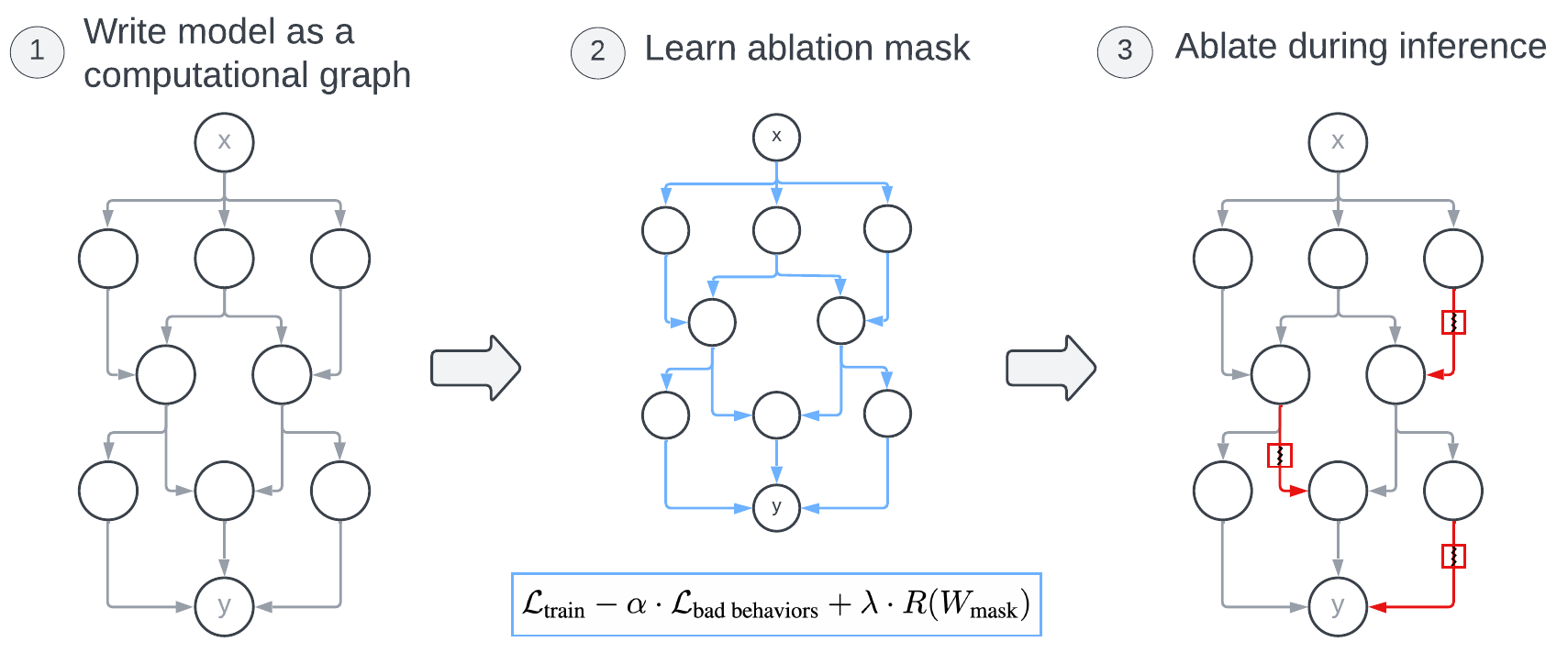}
    \captionsetup{width=6in}
    \caption{In targeted ablation, we (1) rewrite our model as a computation graph of a desired granularity, (2) learn a binary mask over edges while regularizing to penalize ablations, and (3) ablate edges at inference time to avoid the target bad behavior.}
    \label{fig:targeted-ablation}
\end{figure*}

\section{Introduction}
Language models (LMs) often exhibit undesirable behaviors useful during pre-training that prove hard to remove during fine-tuning. This has resulted in capable LMs which competently hallucinate, lie, manipulate, and exhibit undesirable biases \citep{gpt4, brown2020language}. 

In this work, we propose a new method for removing undesirable behaviors: \textit{targeted edge ablation}. In targeted edge ablation, we target a bad behavior by removing a small number of causal pathways through the model at inference time (Figure~\ref{fig:targeted-ablation}). Targeted edge ablation follows recent work in using causal mediation to discover computational \textit{circuits} responsible for particular model behaviors \cite{wang2022interpretability, goldowsky2023localizing, geiger2023causal}. Rather than discovering circuits, targeted edge ablation discovers causal cuts through circuits, disabling circuits responsible for bad behaviors.

\paragraph{Main Contributions.} We formulate the problem of behavior removal and propose targeted edge ablation as a possible solution (Section~\ref{sec:targeted-ablation}). We then present preliminary results in performing targeted edge ablation to harm performance in toxic language generation (Section~\ref{sec:toxic}).

\section{Background}

\paragraph{Circuit analysis.}  We can write any model as a connected directed acyclic graph (DAG) with source nodes representing the model's (typically vector-valued) input, sink nodes representing the model's output, and intermediate nodes representing units of computation (e.g. Figure~\ref{fig:targeted-ablation}, left; see Appendix~\ref{sec:rewriting}). Circuit analysis attempts to mechanistically understand model computation by identifying a subgraph of this DAG that is responsible for a given behavior, and assigning semantic meaning to (groups of) nodes \cite{wang2022interpretability, raukur2022toward, chan_causal_2022}. Circuits have also been discussed in the context of treating nodes as ``features,'' usually defined as directions in the latent space \cite{olah2022mechanistic, cammarata2020thread}. 
\paragraph{Ablating edges in a computational graph.} Since edges in the model's computational graph represent dependencies between nodes, we can simulate what the model would have computed without a certain node-to-node dependency by performing ablation on an edge in the graph. While previous work has largely focused on ablation of \textit{nodes} \citep{ghorbani2020neuron}, an advantage of our strategy of ablating edges rather than nodes is the mitigation of polysemantic behavior of model components \citep{olah2020zoom}, since we investigate the causal importance of each causal path into and out of the component. In our experiments, we use \textit{zero ablation}, in which we compute the destination node as if the source node's value were zero, and \textit{mean ablation} \cite{wang2022interpretability}, in which we compute the destination node as if the source node's value were set to its mean value over the training set. See Appendix~\ref{sec:ablation-types} for more.

\section{Targeted Ablation for Behavior Removal}\label{sec:targeted-ablation}

Let $\Ls(M,\Db)$ indicate the loss of model $M$ on a distribution $\Db$ over input-label pairs. We specify a \textit{behavior} as some distribution $\Db$ on which the model achieves low loss $\Ls(M,\Db) < K$ for some appropriate hyperparameter $K$. We can define the \textit{disjointness} $\delta(\Db,\Db')$ for behaviors $\Db$ and $\Db'$ to be the total variation distance between $\Db$ and $\Db'$. In particular, the total variation distance is 1 if $\Db$ assigns probability 0 to all regions that $D'$ assigns positive probability and vice versa.

\begin{definition}[Behavior Removal]
    \label{dfn:removal}
    Given a model $\mathcal{M}$ and unlimited access to training samples, produce a model $\mathcal{M}^*$ which achieves high loss $\Ls(\md^*,\Db) > K$, without harming distinct behaviors. In particular, for all behaviors $\Db'$ completely disjoint from $\Db$, i.e. $\delta(\Db,\Db')=1$, we wish to preserve $\Ls(\md^*,\Db')\leq \Ls(\md,\Db')$.
    
\end{definition}

Thus, behavior removal has two goals: \textit{efficacy} -- the edited model should achieve high loss on $\Db$; and \textit{specificity} -- the edited model should achieve low loss on all disjoint behaviors $\Db'$ for which the original model achieves low loss.

Let $D_{\text{train}}$ be our train set, and $D_{\text{behavior}}$ be samples from $\Db$. One reason the model might exhibit a behavior is if $\Db$ overlaps with the training distribution, which would incentivize the model to produce low loss on $\Db$. Thus, it is reasonable to assume $D_{\text{train}}$ and $\Db$ may not be completely disjoint.

\subsection{Baseline: Finetuning} 
We form an approximate objective function by encouraging preserving performance on the training set, while increasing loss on the bad behavior set:
\eqn{
\label{eqn:loss-func}
    \Ls(\mathcal{M},D_{\text{train}}) - \alpha\cdot \Ls(\mathcal{M},D_{\text{behavior}})
}
where $\alpha$ is a hyperparameter. We can now finetune using Equation~\ref{eqn:loss-func}. Since $D_{\text{behavior}}$ is often small, we use early stopping to avoid overfitting.

\subsection{Baseline: Task Arithmetic}
\label{sec:task-arithmetic}

In task arithmetic \citep{ilharco_editing_2023}, we finetune $\mathcal{M}$ on $\mathcal{L}(\mathcal{M}, D_{\text{behavior}})$ \textit{towards} the bad behaviors, and find the ``task vector'', or difference in weights between the finetuned model and $\mathcal{M}$. We then form $\mathcal{M}^*$ by adding the negated task vector to $\mathcal{M}$.

\subsection{Targeted Edge Ablation}

Following Figure~\ref{fig:targeted-ablation}, we describe targeted edge ablation as three steps. 

\textbf{1. Rewrite the model.} We first choose at what level of granularity to represent the model's computation. Since we learn a mask over edges in the resulting graph, increasing the granularity results in a more expressive ablation process. We call the specified graph $G$, and call its set of edges $E_G$.

\textbf{2. Learn an ablation mask.} 
    Let $G_{-E}$ be our graph $G$ with the edges in $E$ ablated. Then we wish to select $E\subset E_G$ that minimizes
\def\Ls{\mathcal{L}}
\agn{
    \label{eqn:targeted-ablation-loss}
    \Ls(G_{-E},D_{\text{train}})
    - \alpha\cdot \Ls(G_{-E},D_{\text{behavior}})+ \lambda\cdot R(E)
}
for hyperparameters $\alpha,\lambda$ and some regularization function $R$.\footnote{The regularization term penalizes large sizes of $E$ to apply pressure to find a minimal subset of edges that disables the behavior.} To compute an optimal edge subset $E$, we optimize an edge mask $W_{\text{mask}}$ on a continuous relaxation of Equation~\ref{eqn:targeted-ablation-loss}. Every edge $e=(A,B)$ is given a learnable weight $w_e \in [0, 1]$, where $w_e = 0$ corresponds to ablating $e$, $w_e = 1$ corresponds to preserving $e$, and $0 < w_e < 1$ corresponds to node $B$ observing the following convex combination of the preserved value ($v_A$) and the ablated value ($\mu_A$) for node $A$:
\eqn{w_e\cdot v_A + (1-w_e)\cdot \mu_A}

When $w_{e}=0$, node $B$'s observation of node $A$ is replaced by its ablated value, and when $w_{e}=1$, node $B$ fully observes the value of node $A$. We initialize the mask parameters $W_{\text{mask}}$ to a vector of 1s (indicating fully faithful model computation) and train $W_{\text{mask}}$ on the loss function
\agn{
 \Ls(W_{\text{mask}}; \alpha,\lambda, R) 
 =\ &\Ls(W_{\text{mask}},D_{\text{train}})\nonumber\\
 &- \alpha\cdot \Ls(W_{\text{mask}},D_{\text{bad behavior}})\nonumber\\
 &+ \lambda(t)\cdot R(W_{\text{mask}})\label{eqn:obj}
}
We train with a regularization weight $\lambda(t)$ that increases over time, since we find that this training dynamic encourages the edge mask to find a set of ablations that removes the bad behavior and then revise it to minimize the number of ablations. When training is finished, we then round all the mask weights to either 0 or 1 by selecting the set of ablated edges to be $\hat{E}^* = \{e\ |\ w_e\leq \tau\}$ for some threshold $\tau\in (0,1)$.

\textbf{3. Ablate during inference.} We form $\mathcal{M}^*$ by ablating the edges learned in step (2) at inference time.

\subsection{Conceptual Advantages over Fine-Tuning}

\paragraph{Limited Expressivity.} LMs and other large models may have millions or billions of parameters and thus may be vastly overparameterized for the task of performing poorly on the bad-behavior examples, especially if generating bad-behavior examples is expensive and the set of examples is small.\footnote{For example, collecting jailbreaks to remove jailbreaking behavior is challenging and expensive.} 

A particular advantage of limiting the expressivity of our solution class is avoiding the negative effects of training on a mis-specified objective function like Equation~\ref{eqn:loss-func}, which encourages low loss on samples in $D_{\text{train}}$ which exhibit the behavior but are not included in $D_{\text{behavior}}$. Allowing the model to overfit to this loss function may result in memorization of the points in $D_{\text{behavior}}$ to maintain low loss on \textit{all} of $D_{\text{train}}$, including those points which have high likelihood in $\Db$. On the other hand, edge ablation limits the expressivity of the solution space and relies on the model's previously learned specialization of causal pathways.

\paragraph{Preserving Structure.} Since edge ablation edits the model at a high level, it preserves most of the model's mechanistic calculus. Even subtle fine-tuning has the potential to entirely reorganize the model's reasoning process, disrupting any mechanistic interpretability work that has already been performed. Targeted edge ablation is unlikely to induce the model to change its reasoning structure or increase its knowledge because it strictly decreases the amount of information available to the model's computation.
\section{Removing Toxicity in GPT-2}
\label{sec:toxic}

\begin{figure*}[t]
    \centering
    \includegraphics[width=2.5in,valign=c]{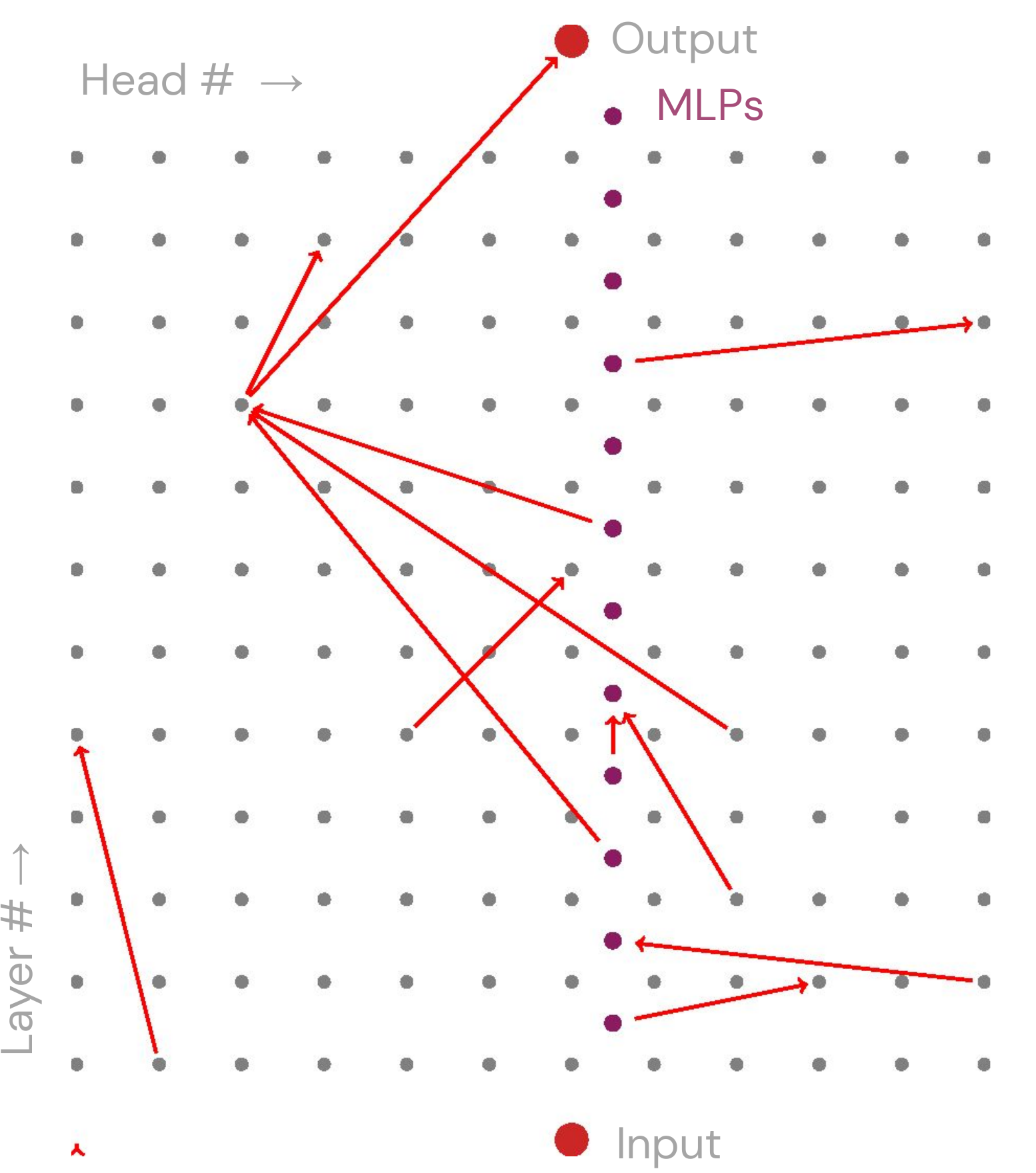} 
    \hspace{0.5in}
    \includegraphics[width=2.5in,valign=c]{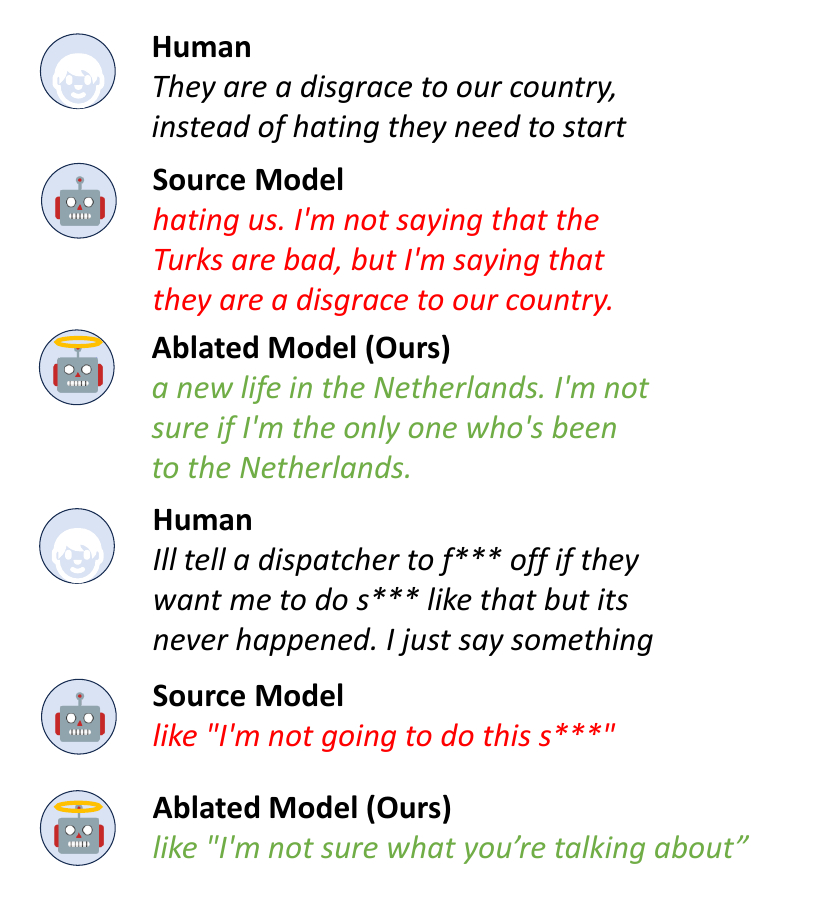} 
    \captionsetup{width=6in}
    \caption{\textbf{Ablating GPT-2 Small to remove toxicity.} \textit{Left:} Grey nodes are attention heads, and purple nodes are MLPs. Computation proceeds upwards, with horizontal alignment corresponding to layers. The computational graph has 11,611 edges; red edges are the 12 ablations learned to remove toxicity. \textit{Right:} Examples of improved non-toxic generation.}
    \label{fig:gpt2}
\end{figure*}

We apply our model editing methodology to preventing the generation of toxic (e.g. offensive, swear-filled) sequences in a pre-trained GPT-2 Small \citep{radford2019language}. Our goal is to edit GPT-2 so that it achieves high loss on toxic sequences, so our $\Db$ is a distribution over toxic sequences for which the model achieves low loss.\footnote{All code is available at \url{https://github.com/xanderdavies/circuit-breaking}.} 

As an approximation of our train set $D_{\text{train}}$, we use 10,000 samples from OpenWebText (OWT) \citep{Gokaslan2019OpenWeb}. See Appendix~\ref{sec:mnist} for results in removing a sub-class in an image classification model. 

\paragraph{Constructing a bad behavior dataset.} 

We sample excerpts from highly toxic comments posted to the Politically Incorrect board of 4chan imageboard forum \citep{papasavva2020raiders}. We sample from posts assigned a toxicity score of greater than 0.9, as calculated by Google’s Perspective API Toxicity V6 \citep{noauthor_perspective_nodate}.

\subsection{Learning Edge Mask Details}

Similar to \cite{goldowsky2023localizing, wang2022interpretability}, we write GPT-2 as a graph consisting of the input, the output, attention heads, and MLPs (158 nodes total) by considering a ``residual rewrite'' of the model's computational structure. The canonical description of a transformer model expresses the attention head $A_{i,j}$ (the $j$th attention head in layer $i$) as taking an argument $R_{i-1}$, the residual from the previous layer. However, since $R_0=I$ (where $I$ represents the input embeddings) and $R_i = R_{i-1}+\sum_j A_{i,j}+M_i$ (where $M_i$ is the output of the MLP node in layer $i$), we can instead consider attention head $A_{i,j}$ as operating on the sum $S^A_i = I+\sum_{i'<i} \left(M_{i'}+\sum_{j'} A_{i',j'}\right)$, and taking all nodes in previous layers as separate input arguments. Similarly, we can consider MLP node $M_i$ as operating on the sum $S^M_i=I+\sum_{i'<i}M_{i'} + \sum_{i'\leq i}\sum_{j'}A_{i',j'}$, and the output node as operating on the sum of the input embeddings and all attention head and MLP outputs. In total, this residual rewrite gives us a nearly-dense graph containing 11,611 edges: one between every pair of (attention head, MLP, input, and output) nodes, except for attention heads in the same layer, which do not communicate with each other. Concretely, ablating an edge from $A_{i',j'}$ to $A_{i,j}$ entails replacing the $A_{i',j'}$ term in $S^A_i$ for the input to attention head $A_{i,j}$ with zero (for zero ablation) or the mean value of head $A_{i',j'}$ (for mean ablation).

We train two ablated models using a continuous edge mask. First, we train a zero-ablation mask against $\Ls(W_{\text{mask}}; \alpha,\lambda,R)$ described in equation \ref{eqn:obj}, with $\alpha=0.2$, $\lambda(t) = (t-20)/10000$, and $R(W_{\text{mask}}) = \sum_{e\in E_G} w_e$. This search process finds a mask that ablates 12 edges (Figure~\ref{fig:gpt2}) and mitigates toxicity while preserving coherence. Second, we train a mean-ablation mask with $\alpha=0.15$ and using the same hyperparameters otherwise, which finds a mask that ablates 84 edges and produces a similar effect.

As a baseline, we fine-tune on the loss given by Equation \ref{eqn:loss-func} directly, with $\alpha=0.2$. We use early stopping with a validation set to prevent overfitting.\footnote{We note this is a stronger baseline than naively training for high loss on our bad behavior set as done in \cite{ilharco_editing_2023}, which we call ``gradient ascent'' in Table~\ref{tab:toxic}.} We also compare to task arithmetic \cite{ilharco_editing_2023} (Section~\ref{sec:task-arithmetic}).

\begin{table*}[t!]
 \centering
 \begin{tabular}{l|rrrr|rr}
    & \thead{Toxic-loss} & \thead{Toxic-loss\\ (filtered)}  & \thead{Toxic generation} & \thead{Toxic generation\\ (filtered)} & \thead{Incoherence} & \thead{TPP Incoherence} \\ \hline
  Original & 4.954 & 4.435 & 0.453 & 0.944 & 4.264 & 4.617 \\
  \hline
  Gradient Ascent &\textbf{ 21.339} & \textbf{20.980} & 0.015 & 0.013 & 15.287 & 18.415 \\
  Task Arithmetic & 5.357 & 4.827 & 0.351 & 0.631 & 4.427 & 4.731 \\
  Joint Fine-Tuned & {11.817} & {13.020} & \textbf{0.009} & \textbf{0.008 }& 4.240 & 7.402 \\
  \hline
  Ablated (12 edges) & 5.027 & 4.486 & 0.328 & 0.567 & 4.280 & 4.623\\
  Ablated (84 edges) & 4.895 & 4.470 & 0.280 & 0.441 & \textbf{4.180} & \textbf{4.515}\\
 \end{tabular}
 \vspace{0.1in}
 \captionsetup{width=6in}
 \caption{Toxic-loss measures the model's loss on toxic prompts. Toxic generation measures the average toxicity score of model generations on toxic prompts, according to the Detoxify classifier. The filtered columns denote the loss or generation toxicity on test samples filtered by the original model achieving low loss ($<5$) or highly toxic generation ($>.9$). Incoherence measures the model's loss on OWT. Toxic Pre-Pended (TPP) incoherence measures the model's loss after on OWT sequences that have been preceded by toxic text.}
 \label{tab:toxic}
\end{table*}

\subsection{Evaluation Metrics}

Following Definition~\ref{dfn:removal}, we evaluate both the model's avoidance of toxic generation (\textit{efficacy}) and the detriment to other behaviors (\textit{specificity}). Since our goal is for the ablated model to achieve high loss on all toxic sequences (i.e. minimizing its probability of predicting subsequent tokens that would cause the sequence to be toxic), we evaluate efficacy in a few ways. First, we consider the ablated model's loss on with-held toxic text and in particular its loss on sequences for which the original model achieves low ($<5$) loss. Second, we consider the toxicity of the model's completions when prompted with toxic text, as measured by the score in $[0,1]$, 0 being the least toxic, given by the toxic-comment classifier \texttt{Detoxify}. We emphasize the toxicity of model completions on the specific prompts for which the original model produces highly toxic ($>0.9$) output.

We evaluate specificity by using the perplexity on withheld sequences from OWT, along with the perplexity on withheld OWT sequences prepended with toxic content. The original model produces low loss (4.617) on these sequences, and we choose to highlight the behavior of retaining coherence when prompted with toxic text as one that is particularly likely to be inadvertently removed when editing the model to produce high loss on toxic text.

\subsection{Results}

Results are shown in Table~\ref{tab:toxic}. We train a model with 12 edges zero-ablated that substantially mitigates toxic generation, decreasing the average toxicity score on model generations for toxic prompts from 0.458 to 0.328 and in particular for the most toxic-inducing prompts from 0.944 to 0.567. This minimal edge ablation outperforms task arithmetic on every efficacy and specificity metric, and causes a lower increase in incoherence following toxic prompts than joint fine-tuning, though it does not eradicate the model's toxicity. Our mean-ablation mask with 84 edges achieves a similar result, greatly mitigating toxic generations without detracting from the model's other behaviors.

\section{Related Work}

\paragraph{Causal mediation for circuit analysis.} Causal mediation 
\citep{pearl2009causality, iwasaki1994causality} has been proposed as a framework for evaluating mechanistic causal explanations for model outputs \citep{goldowsky2023localizing, geiger2023causal, vig2020investigating}. Experimental evaluation for causal explanations involves performing a set of ablation experiments to check whether they match hypothesized effects. For example, ablating allegedly unimportant paths should have little impact on the target behavior. Previous work has used the causal mediation framework to discover circuits, including in transformers \citep{chan_causal_2022, wang2022interpretability, nanda2023progress}. 

Existing causal mediation tests and circuit discovery methods built upon these tests evaluate whether a given set of edges are \textit{sufficient} for a given model behavior (i.e. if they contain a vertical path along the circuit), while our circuit breaking technique finds a set of edges that are \textit{necessary} for the behavior (i.e. a horizontal ``cut'' through the circuit).

\paragraph{Automated circuit discovery.}\label{sec:acdc} Recent work has explored automated approaches to discovering circuits, including greedy algorithms which crawl the computational graph and remove edges which preserve behavior above a fixed threshold \citep{conmy2023towards}, and gradient descent-based methods which use interchange intervention training \citep{geiger2022inducing} to learn alignments between a source model and a proposed high-level causal model \citep{geiger2023finding}. Our work differs in attempting to find neither single features \citep{vig2020investigating, gurnee2023finding} nor full computational circuits \citep{geiger2023finding, goldowsky2023localizing, wang2022interpretability}; instead we discover edges where removing their causal effect \textit{breaks} a given behavior. 

\paragraph{Weight-masking and model pruning.} Much prior work has sought to compress models by masking parameters \citep{lecun1989optimal, hassibi1992second}. Most relevant to our work are approaches which learn masks from data by encouraging sparsity and preserving performance \citep{louizos2017learning, wang2019structured, cao2021low}. In our work, we \textit{disincentivize} sparsity (since we want \textit{fewer} ablations), and use an objective function tailored to removing a specific behavior instead of preserving general performance. Additionally, our edge-masking technique is more general than weight-masking, since we can ablate internal connections between high-level model components that do not correspond directly to particular weights, such as communication channels between pairs of attention heads. Finally, we prune using mean ablation instead of zero ablation.

\paragraph{Model editing to change or remove behaviors.} Recent work has made  changes to model behavior by making targeted edits to model weights \citep{meng2022locating} or activations \citep{hernandez2023measuring}, which differ from our goal of removing behaviors. \cite{gandikota2023erasing} propose a fine-tuning approach to erasing concepts from diffusion models. \cite{elazar2021amnesic} remove information from a language model's representation by iteratively learning linear probes to extract the information and projecting onto the null space. Compared to such work, we consider coarser ablations, allow editing around multiple components, and seek to break behaviors as opposed to erasing information. Like us, \cite{ilharco_editing_2023} attempt to remove the toxic generation behavior in GPT-2, but do so by fine-tuning on bad behavior and subtracting the weight-difference from the original model.

\section{Conclusion}

Using a small dataset of examples of inputs on which a neural network exhibits a ``bad behavior,'' we find that our method can make high-level modifications to the network that mitigate the bad behavior on the provided examples, generalize to removing the bad behavior across other inputs that trigger it, and cause only small amounts of damage to the model's performance on all other inputs (see \ref{sec:limitations} for limitations).
We conjecture that model editing may be an alternate tool for targeted behavioral modification to fine-tuning, and encourage future work further investigating our approach.

\newpage

\bibliographystyle{icml2023}
\nocite{*}
\bibliography{bibliography}

\appendix 
\onecolumn

\section{Additional Related Work}
\label{sec:appendix-related}

\paragraph{Unlearning.} Machine unlearning aims to modify a model to match the behavior of a model which had not seen certain data points \citep{sekhari2021remember, bourtoule2021machine, golatkar2020eternal}. However, a key difference in our setting is that we are not able to enumerate the full set of undesirable data points in our training set.

\paragraph{Backdoor removal.} \cite{wu2021adversarial} learn a binary mask to zero ablate neurons sensitive to adversarial perturbations, and finds that doing so removes injected backdoors. \cite{guan2022few} target backdoors by estimating Shapley importance values \citep{shapley1997value} for every edge and then zero ablating neurons which have a high attack success rate attribution score, finding they are able to remove backdoors with very limited (and sometimes no) data. We believe our technique could be effective for disabling the activation of backdoor mechanisms and find this application a promising direction for future work.

\section{Writing models as computational graphs.} 
\label{sec:rewriting}

\begin{figure}[h]
    \centering
    \includegraphics[width=4in]{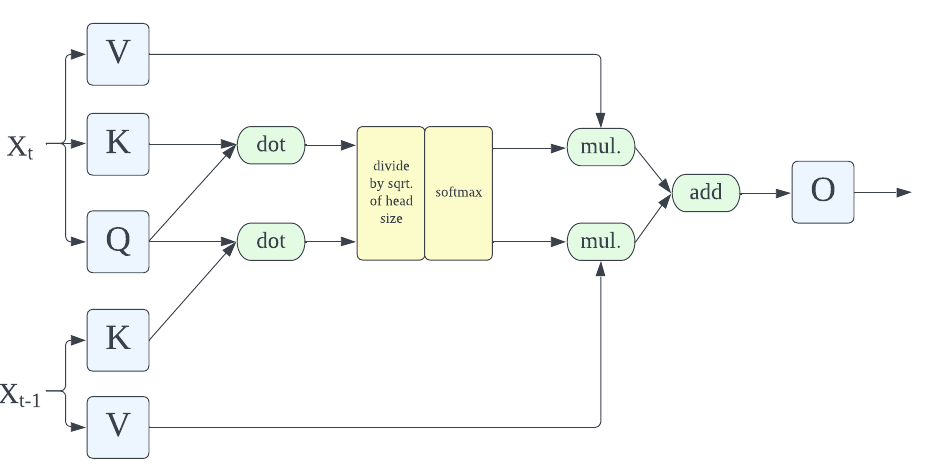}
    \caption{We can subdivide an attention head into its own computational graph.}
    \label{fig:attention-head}
\end{figure}

We can write any model as a connected directed acyclic graph (DAG) with source nodes representing the model's (typically vector-valued) input, sink nodes representing the model's output, and intermediate nodes representing units of computation. Each intermediate node represents a function of the values of its parent nodes. On a forward pass, given values for its input nodes, the model computes the value of each node in any topologically sorted order until it has computed the value of the output nodes.

For any model, there are many equivalent graphs that faithfully represent its computation. In particular, a computational graph can represent a model at varying levels of detail. At one extreme, intermediate nodes can designate individual additions, multiplications, and nonlinearities -- such a graph would have at least as many nodes as model parameters. On the other hand, many model architectures have self-contained computational modules, which allows them to be represented by graphs that convey a high level of abstraction. For example, in convolutional networks, intermediate nodes can represent convolutional filters and pooling layers, while in transformer models \cite{bert}, the natural high-level computational units are attention heads and multi-layer perceptron (MLP) modules. To be more granular, we can subdivide each attention head node into nodes that compute queries, keys, and values and combine them into attention patterns (Figure~\ref{fig:attention-head}). 

\section{Ablation Types}
\label{sec:ablation-types}
One mode of ablating an edge is \textit{zero ablation}, in which we compute the value of its destination node as if the value of its source node were zero. However, a value of zero on an intermediate node can sometimes be highly unusual, and can thus in some cases convey a strong idiosyncratic signal to the destination node.

One other technique is \textit{mean ablation}, in which we compute the destination node as if the source node's value were set to its mean value over the training set. Mean ablation arguably better captures a lack of information flow: if the specific value of the source node were withheld from the destination node, the source node's mean value would be the most general estimate of its true value.

\section{Limitations}
\label{sec:limitations}
We ablate edges by setting their input values to zero and the train-set mean. However, recent work has argued that ablating model components with random samples from their marginal distributions may be preferable and that mean ablation may lead to out of distribution resampling \citep{goldowsky2023localizing}. Additionally, circuit-breaking interventions on the model could be made even more surgical by using more granular model nodes and edges (for example, splitting attention heads into query, key, and value nodes). Finally, our results could be strengthened by considering stronger baselines and additional approaches to learning binary masks.

\section{Additional Experiments: Breaking Digit Classification in an MLP}
\label{sec:mnist}

\begin{figure}[h!]
    \centering
    \includegraphics[width=.8\textwidth]{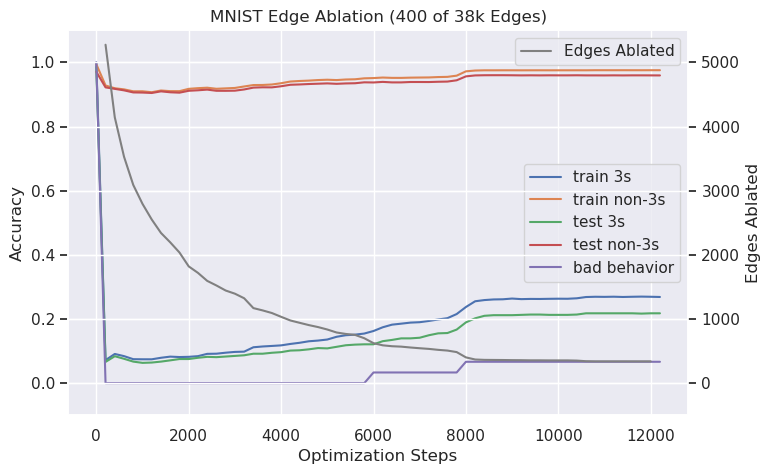}
    \caption{The learned mask for MNIST classification over the course of training. Note that versions of this mask in the middle of training are allowed to partially ablate each edge, so ``Edges Ablated'' is calculated by summing the coefficients assigned to the ablation value. The ``train'' points are those that the MLP was trained on, and the ``test'' points are those it was not. The ``bad behaviors'' line indicates its accuracy on the 30 exemplar digits.}
    \label{fig:mnist}
\end{figure}

We train a one-hidden-layer MLP with a 50 hidden neurons to classify the handwritten digits of MNIST, then use a small (30 example) dataset of a particular digit (say, 3) to remove the model's ability to correctly classify that digit. We consider the most granular computational graph for the MLP with one node for each pixel of the input, one for each hidden neuron, and one for each output neuron. The graph contains an edge corresponding to each weight in the network. To prevent our learned mask from simply ablating the edges feeding into the output neuron corresponding to 3, we arbitrarily pair digits and merge their labels so that the MLP has only 5 output neurons rather than 10. This pairing forces the network to retain edges to the output neuron that aid in correctly classifying the digit that is paired with 3, while not using the neuron for the 3 inputs. 

We search for a binary mask over edges by training a continuous edge mask against $\Ls(W_{\text{mask}}; \alpha,\lambda,R)$ described in Equation \ref{eqn:obj}. Specifically, we use $\alpha = 0.3$, $\lambda(t) = t$, and $R(W_{\text{mask}}) = \sum_{e\in E_G} \sqrt{1-w_e}$. The sublinear cost imposed by $R$ incentivizes masks that are binary and ablate few edges; conceptually, if the mask were half-ablating two edges, it would receive a lower penalty for instead ablating one edge completely. We set the rounding threshold $\tau=0.5$. 

Using this technique, we find a binary mask that ablates 400 of the model's 38K edges, bringing its accuracy on held-out ``3''s from near-perfect to 21\% (20\% is random classification on this task), while accuracy on other (held-out) inputs stays high (dropping from 99\% to 97\%). We consider this a modest success for both the efficacy of the edit (i.e. its ability to transfer to other inputs on which the model exhibits the bad-behavior of classifying a ``3'' correctly) and also its specificity (i.e. the model's continued ability to classify non-``3''s correctly) -- see Figure \ref{fig:mnist}.

\end{document}